\pdfoutput=1
\documentclass[11pt]{article}

\usepackage{ACL2023}

\usepackage{times}
\usepackage{latexsym}

\usepackage[T1]{fontenc}
\usepackage[utf8]{inputenc}

\usepackage{microtype}

\usepackage{inconsolata}

\usepackage{mathtools}

\usepackage{array}
\usepackage{geometry}
\usepackage{float}

\title{Personality Differences Drive Conversational Dynamics: A High-Dimensional NLP Approach}

\author{Julia R. Fischer \and Nilam Ram \\
  Stanford University \\
  \texttt{\{juliafischer, nilamram\}@stanford.edu}}

\begin{document}
\maketitle
\begin{abstract}
This paper investigates how the topical flow of dyadic conversations emerges over time and how differences in interlocutors’ personality traits contribute to this topical flow. Leveraging text embeddings, we map the trajectories of $N = 1655$ conversations between strangers into a high-dimensional space. Using nonlinear projections and clustering, we then identify when each interlocutor enters and exits various topics. Differences in conversational flow are quantified via \textit{topic entropy}, a summary measure of the "spread" of topics covered during a conversation, and \textit{linguistic alignment}, a time-varying measure of the cosine similarity between interlocutors' embeddings. Our findings suggest that interlocutors with a larger difference in the personality dimension of openness influence each other to spend more time discussing a wider range of topics and that interlocutors with a larger difference in extraversion experience a larger decrease in linguistic alignment throughout their conversation. We also examine how participants’ affect (emotion) changes from before to after a conversation, finding that a larger difference in extraversion predicts a larger difference in affect change and that a greater topic entropy predicts a larger affect increase. This work demonstrates how communication research can be advanced through the use of high-dimensional NLP methods and identifies personality difference as an important driver of social influence.

\end{abstract}

\section{Introduction}
Each conversation we take part in is an opportunity for others to influence us and for us to influence others. Among other categorizations, this social influence can be cognitive, such as introducing someone to a new topic to think about, or affective (related to emotion), such as making someone feel more positive by the end of a conversation as compared to before. However, not every conversation will induce changes in interlocutors' cognition or affect. How can we determine which ones will? We might consider \textit{whom} one talks to as an important predictor of the conversation's resulting social influence. Aspects of an interlocutor's personality are reflected in the language they use \citep{pennebaker1999}. Thus, conversation can serve as a mechanism through which differences in personality drive social influence.

Large conversational corpora and the computational tools developed for working with linguistic data open new opportunity to test theories of social influence at a large scale. Identifying the mechanisms and drivers of social influence in real-world data is necessary for furthering our basic understanding of how conversations serve as vehicles for cognitive and affective change. This foundational knowledge can help inform downstream theories of applied social influence tasks, such as negotiation \citep{distasi2024, glenn2010} and persuasion \citep{huma2020, wood2000}.

In this paper, we investigate how differences in interlocutors' personality traits relate to the content and outcomes of a conversation. We focus on how personality differences relate to the topical flow of a conversation, the linguistic alignment between its interlocutors, and the interlocutors' subjective ratings of their affect. Through this analysis, we set forth a method of characterizing conversational behavior through high-dimensional text embeddings, projection into a low-dimensional space, and topic clustering.

\section{Related Work}

\subsection{Conversation Analysis}

Traditional methods of analyzing conversation are primarily qualitative and focus on the manual identification of several instances of a conversational phenomenon \citep{hoey2017, silverman2006}. While this approach generates rich data about a specific conversational behavior, it requires the phenomenon of interest to be clearly defined in advance. The study of the broader topics that emerge during a conversation has largely been done through traditional conversation analysis \citep{todd2011, yang2019}. Recent advancements in natural language processing (NLP) tools makes possible the automated identification and quantitative analysis of conversation topics \citep{wallach2006}.

Understanding social influence phenomena requires not only identifying conversation topics, but also how interlocutors move among them over time. For example, two interlocutors may begin a conversation in starkly different topics, then become more synchronized in the topics they visit toward the end of the conversation. This could indicate that some form of social influence occurred during the conversation to bring the two into closer alignment. Temporal analysis of conversation topics can be facilitated by defining a topic “space” within which interlocutors move \citep{templeton2024}. High-dimensional text embeddings have been used to project conversational turns into a semantic space \citep{onell2024, vakulenko2018}, which can then be used to track the topical flow of a conversation over time. 

\subsection{Personality in Dialogue Systems}

Although our work focuses on human-human conversations, personality is relevant in other kinds of interactions, including those with automated dialogue systems, and how they facilitate social influence. In human-machine interactions, a machine system can detect a human’s personality traits through analysis of their conversational behavior \citep{ivanov2011, mairesse2007}. Machine agents can also express personality traits through the language they generate, as detected by human interlocutors \citep{mairesse2008}. Thus, it is plausible that personality functions in human-machine conversations similarly to how it does in human-human conversations.

With their generativity and flexibility, large language models (LLMs) are especially capable of adopting personality traits. In fact, LLMs may already exhibit particular traits without needing to be prompted. \citet{hilliard2024} demonstrate that LLMs generally exhibit high openness and low extraversion and that newer models with more parameters exhibit a broader range of personality traits. Thus, the degree of social influence in a human-machine conversation may vary based on the model with which one interacts. LLMs also show promise for adopting personality traits through prompting, then displaying these traits through their text outputs \citep{jiang2023, serapio-garcia2023}. However, when an LLM agent interacts with another LLM agent, it may struggle to maintain a consistent personality, instead aligning to produce utterances similar to that of the agent they are conversing with \citep{frisch2024}. A better understanding of personality's role in human-human conversations may help us strengthen the social influence capabilities of LLMs.

\section{Data}

\subsection{Overview}

To investigate conversational dynamics in relation to interlocutors' personality traits, we make use of the CANDOR (Conversation: A Naturalistic Dataset of Online Recordings) corpus, collected by \citet{reece2023}. The corpus comprises 1656 dyadic conversations that were facilitated through online video chat in the year 2020. The corpus provides multimodal data on a rich set of unscripted, naturalistic conversations in which interlocutors influence and respond to each other without specific constraints or goals. 

\subsection{Participants \& Procedure}

Participants were 1456 unique individuals ages 19 to 66 years who were located all across the United States  \citep{reece2023}. In brief, participants were matched, based on scheduling availability (without using any demographic information), with other participants to have human-human dyadic conversations online. All participants consented to having their conversation's audio and video recorded and released for research purposes. At the scheduled time, pairs of participants joined a video meeting and chatted with their conversation partner for at least 25 minutes. They were not given any specific guidelines about what to discuss. Each member of the dyad was compensated up to \$15 for participating in the recorded conversation and completing pre- and post-conversation surveys. Although more than half of the participants engaged in multiple conversations, all 1656 conversations in the corpus were obtained from unique dyad pairs. Our analysis is based on 1655 conversations, after removing one conversation that contained a very small number of utterances. For each conversation in the CANDOR corpus, there are two main data components: the transcript and the survey. 

\subsection{Transcriptions}

The transcript data consists of turn-by-turn transcriptions of each conversation. The 850+ hours of recorded conversations were transcribed and parsed into conversational turns using three different turn segmentation algorithms that differ in how they track when the floor is passed back and forth from one interlocutor to the other: Audiophile, Cliffhanger, and Backbiter \citep{reece2023}. In brief, Audiophile, the most basic algorithm applied to the data, initiates a new turn each time an interlocutor starts speaking. In contrast, the Cliffhanger algorithm ends the current turn and starts a new turn when the interlocutor reaches a terminal punctuation mark (i.e., a period, exclamation point, or question mark). Thus, whereas the Audiophile algorithm passes the floor back and forth whenever an interlocutor uses a backchannel acknowledgement like "mm-hmm," the Cliffhanger algorithm embeds backchannels within more substantive utterances. Pushing the conceptual meaning of backchannel utterances further---as meaningful utterances that can signal affiliation and understanding---the Backbiter algorithm identifies backchannel responses and separates them from the main transcript into a separate backchannel transcript. Thus, Backbiter produces two transcripts (main and backchannel) that run in parallel. An excerpt from one conversation, as segmented into turns by Audiophile, Cliffhanger, and Backbiter, respectively, is shown in Table \ref{table:Table 1}.

Following our interest in parsing conversational flow, we sought a transcript structure that captured the conversational moves made by each interlocutor as they moved through different topics. As noted by \citet{reece2023}, the Audiophile algorithm provides a rather aggressive division of turns. As can be seen in Table \ref{table:Table 1}, all instances in which the current non-speaking interlocutor uses a backchannel acknowledgement like "okay" or "yeah" are considered new speaking turns. Thus, Audiophile-based transcripts tend to have many very short speaking turns that do not include topical words and also fracture interlocutors' conveyance of thoughts or topics across multiple turns. In contrast, the Backbiter algorithm completely separates the backchannel turns from the conversational flow. We thus chose to analyze conversational flow using the transcripts produced by the Cliffhanger algorithm, as these transcripts provided smoother coverage of the topics engaged during each conversation (e.g., via longer turns) while retaining some of the social influence (e.g., rapport) provided through backchannel utterances.

\begin{table*}
\centering
\begin{tabular}{|p{0.31\textwidth}|p{0.31\textwidth}|p{0.31\textwidth}|}
\hline
\textbf{Audiophile Algorithm} & \textbf{Cliffhanger Algorithm} & \textbf{Backbiter Algorithm} \\
\hline
\textcolor{red}{\textbf{A:}} So are you from like the Chicago area or elsewhere? \newline
\textcolor{blue}{\textbf{B:}} Uh, Chicago is about an hour away from us… \newline
\textcolor{red}{\textbf{A:}} \textbf{Okay.} \newline
\textcolor{blue}{\textbf{B:}} from… \newline
\textcolor{red}{\textbf{A:}} That's cool. \newline
\textcolor{blue}{\textbf{B:}} I don't know what the, not, not, not downstate, but like, you know the mm… near there… \newline
\textcolor{red}{\textbf{A:}} \textbf{Yeah.} \newline
\textcolor{blue}{\textbf{B:}} basically. \newline
\textcolor{red}{\textbf{A:}} \textbf{Yeah. Sure.} \newline
\textcolor{blue}{\textbf{B:}} I've been to Chicago. My dad, um, lived there for like, you know, he grew up there, he met my mom there, you know? &

\textcolor{red}{\textbf{A:}} So are you from like the Chicago area or elsewhere? \newline
\textcolor{blue}{\textbf{B:}} Uh, Chicago is about an hour away from us from… I don't know what the, not, not, not downstate, but like, you know, the mm… near there basically. \newline
\textcolor{red}{\textbf{A:}} \textbf{Okay.} That's cool. \textbf{Yeah. Yeah. Sure.} \newline
\textcolor{blue}{\textbf{B:}} I've been to Chicago. My dad, um, lived there for like, you know, he grew up there, he met my mom there, you know? &

\textcolor{red}{\textbf{A:}} So are you from like the Chicago area or elsewhere? \newline
\textcolor{blue}{\textbf{B:}} Uh, Chicago is about an hour away from us from… \newline
\textcolor{red}{\textbf{A:}} That's cool. \newline
\textcolor{blue}{\textbf{B:}} I don't know what the, not, not, not downstate, but like, you know, the mm… near there basically. I've been to Chicago. My dad, um, lived there for like, you know, he grew up there, he met my mom there, you know?
\\
\hline
\end{tabular}
\caption{Audiophile vs. Cliffhanger vs. Backbiter algorithms' turn segmentation of the same portion of a conversation transcript. Backchannel utterances bolded.}
\label{table:Table 1}
\end{table*}

\subsection{Survey Measures}

Survey data were collected in a three-part process. Participants completed a screening questionnaire when enrolling in the study where they provided basic demographic information. Immediately before each conversation, participants completed a pre-conversation survey where they reported on their current affective state. Then, immediately after each conversation, participants completed a post-conversation survey where they reported on a variety of psychological states, including their current affective state, psychological traits, and their perceptions of the conversation partner. The specific measures used in our analysis are described here. 

\subsubsection{Personality}

Participants' personality traits were measured during the post-conversation survey using the Big Five Inventory. Participants indicated their level of agreement (5-point Likert scale) with 15 statements related to the personality traits of openness, conscientiousness, extraversion, agreeableness, and neuroticism. Personality trait scores for each interlocutor were calculated as the average of the relevant item ratings for each of the five dimensions. 

\subsubsection{Affect and Affect Change} 

Participants' affective states were measured immediately prior to and again after the conversation. In our analysis we specifically make use of responses to the item, "To what extent do you feel positive affect (e.g., good, pleasant, happy) or negative affect (e.g., bad, unpleasant, unhappy) right now?" that were provided on a 9-point scale ranging from "extremely negative" to "extremely positive". In addition to pre-conversation and post-conversation affect valence scores, we computed for each interlocutor in each conversation an \textit{affect change} score as the difference between the post- and pre-conversation scores, where more positive scores indicate larger increases in positive affect.

\section{Method}

Conversational flow and dynamics captured in the Cliffhanger-based transcripts were summarized in a multi-step process that made use of a variety of computational methods and tools. We characterized the flow and dynamics of each conversation by computing several metrics based on both high- and low-dimensional representations of the conversations.

\subsection{Text Embeddings: Mapping Conversation in High-Dimensional Space}
First, we mapped the conversation transcripts into numerical vectors using a standard set of text embeddings that were developed on other corpora, specifically the SentenceTransformers Python framework \citep{reimers2019}. In particular, we used SentenceTransformers' all-mpnet-base-v2 model \citep{song2020} to compute a 768-dimension sentence embedding for each utterance in each conversation. These embeddings thus provide a collection of 768-dimensional dyadic time series that chronicle the turn-by-turn evolution of each of the 1655 conversations---specifically how interlocutors A and B led, followed, and moved with each other through the high-dimensional space. 

\subsubsection{Conversation Metric: Linguistic Alignment across the Conversation}
Using the 768-dimensional vectors, we calculated the time-varying \textit{linguistic alignment}, or the degree of similarity between interlocutors' language, as the cosine similarity \eqref{eq:3} between successive speaking turns: the first embedding representing an utterance from Interlocutor A and the second embedding representing the consecutive utterance from interlocutor B.

\begin{equation}
  \begin{aligned}
    S_c(A, B) &= \cos(\theta) = \frac{\mathbf{A} \cdot \mathbf{B}} {\lVert \mathbf{A} \rVert \lVert \mathbf{B} \rVert} \\
    & = \frac{\sum_{i=1}^{n} A_i B_i}{\sqrt{\sum_{i=1}^{n} A_i^2} \sqrt{\sum_{i=1}^{n} B_i^2}} \label{eq:3}
  \end{aligned}
\end{equation}

We then summarized how linguistic alignment changed across each conversation by modeling the cosine similarity scores as a function of time, specifically, turn in conversation. We ran this regression \eqref{eq:4} for each of the 1655 conversations separately to obtain three summary linguistic alignment (LA) metrics: \textit{LA intercept}, \textit{LA linear change}, and \textit{LA quadratic change}. We elected to include time as a quadratic polynomial predictor to capture potential nonlinearity in linguistic alignment over time. Note that turns are typically shorter near the beginning of a conversation and become longer as the conversation progresses \citep{edwards2024}, so these coefficients should not be interpreted as exactly linear with respect to time.

\begin{equation}
  \begin{aligned}
    CosineSimilarity_t & = \beta_0 + \beta_1 Turn_{t} \\
      & + \beta_2 Turn^2_{t} + \epsilon_{t} \label{eq:4}
  \end{aligned}
\end{equation}

\subsection{Topics: Projection and Clustering Conversation in Low-Dimensional Space}
To identify when interlocutors entered and exited areas of the space that might hold specific and human-interpretable meaning, we used nonlinear projection to cast the locations of each speaking turn in the 768-dimensional embedding space into a two-dimensional space. Specifically, we computed the projection using Uniform Manifold Approximation and Projection (UMAP) \citep{mcinnes2020}, a technique that reduces dimensionality while preserving the global topological structure of the data. The minimum distance parameter was set at 0.2 (above the 0.1 default) so that the two-dimensional projection would be more spread out, and thus facilitate topological separation and identification of conversation topics.

\begin{figure}[H]
    \includegraphics[width=7.5cm]{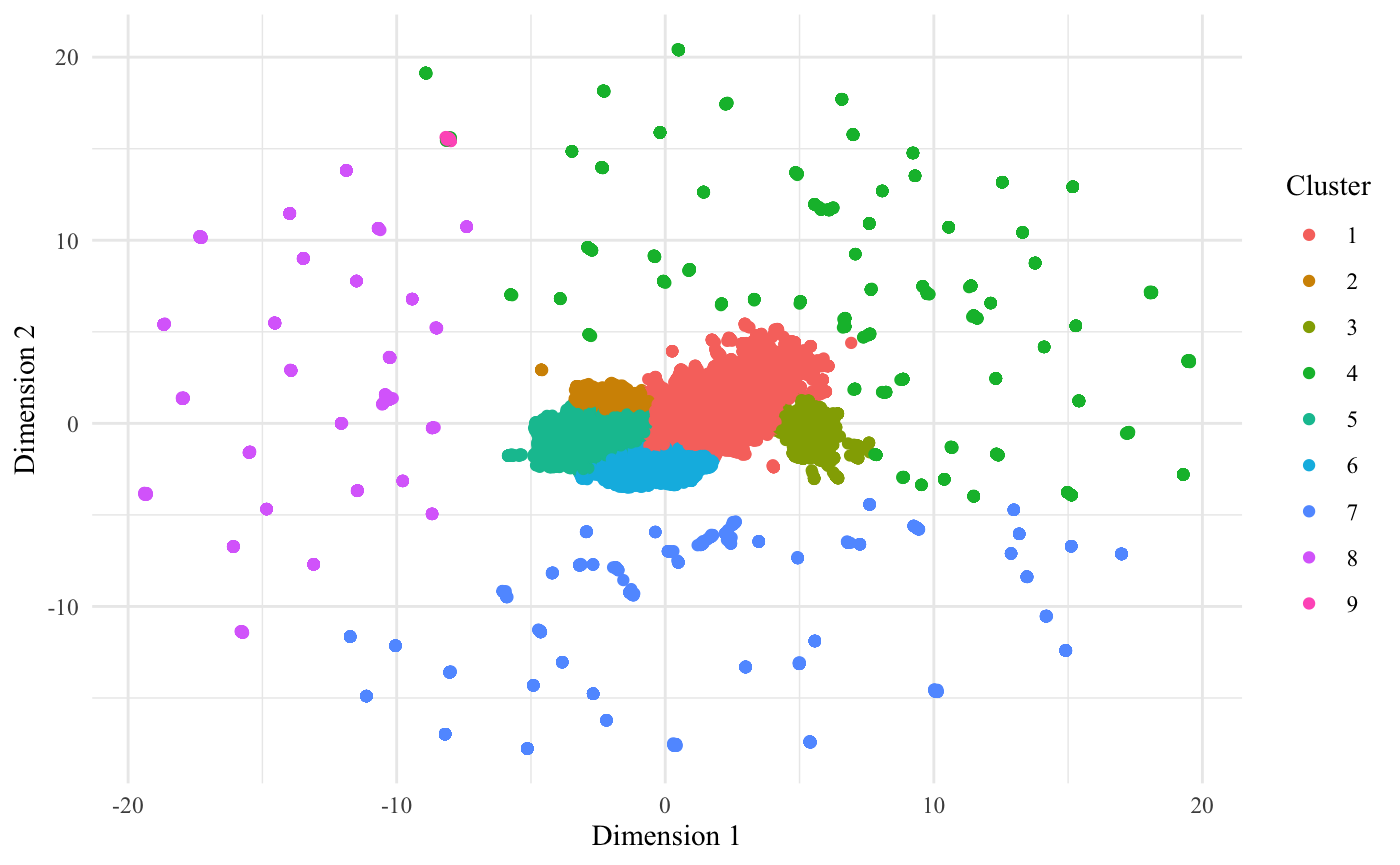}
    \caption{Topic clusters determined by model-based clustering on a thinned set of 10 utterances sampled from each conversation.}
    \label{fig:Figure 1}
\end{figure}

After randomly sampling 10 utterances from each of the 1655 conversations and projecting them into the two-dimensional space obtained via UMAP, we identified discernable areas of the space using cluster analysis. In particular, we used Mclust \citep{scrucca2023}, an R package that fits Gaussian finite mixture models for model-based clustering using an expectation-minimization (EM) algorithm. The optimal number of clusters, selected by minimizing the Bayesian information criterion (BIC) \eqref{eq:1}, was nine. 

\begin{equation}
    BIC = k \ln(n) - 2 \ln(\hat{L}) \label{eq:1}
\end{equation}
Locations of the random subset of speaking turns in the two-dimensional space and their cluster assignments are shown in Figure \ref{fig:Figure 1}. This optimal cluster solution was then used to compute cluster assignments for all speaking turns in all 1655 conversations.  

Face validity of the topic clustering was examined by extracting the top differentiating keywords from each cluster. After filtering out stop words, as well as words occurring at very low and very high frequencies, keywords for each topic were identified as those word stems with the greatest "keyness," a measure of the differential occurrence of a word stem in one topic versus in the remaining topics \citep{benoit2018, bondi2010}. As shown in Table \ref{table:Table 2}, the keywords representative of several of the clusters were interpretable as relatively cohesive topics. Other clusters, however, were less interpretable, "catch-all"-type topics.

\begin{table}[H]
\centering
\begin{tabular}{ccc}
\hline
\textbf{Topic A} & \textbf{Topic B} & \textbf{Topic C} \\ \hline
dog & live & school \\ \hline
mask & citi & survey \\ \hline
cat & famili & class \\ \hline
wear & york & play \\ \hline
vote & california & onlin \\ \hline
trump & move & watch \\ \hline
pet & state & studi \\ \hline
elect & area & money \\ \hline
hair & florida & job \\ \hline
breath & place & prolif \\ \hline
\end{tabular}
\caption{Top 10 keywords for three of the nine topic clusters.}
\label{table:Table 2}
\end{table}

\subsubsection{Conversation Metric: Topic Entropy}

Using the cluster assignments for each speaking turn, we introduce and use \textit{topic entropy} as a summary measure quantifying the "spread" of topics covered during a conversation. Specifically, topic entropy was computed for each conversation as the Shannon entropy \eqref{eq:2} of the cluster assignments of all speaking turns in that conversation. 

\begin{equation}
    H(X) = -\sum_{i=1}^{n} p(x_i) \log_2 p(x_i) \label{eq:2}
\end{equation}

\subsection{Example Conversations}

\begin{figure}
    \includegraphics[width=7.5cm]{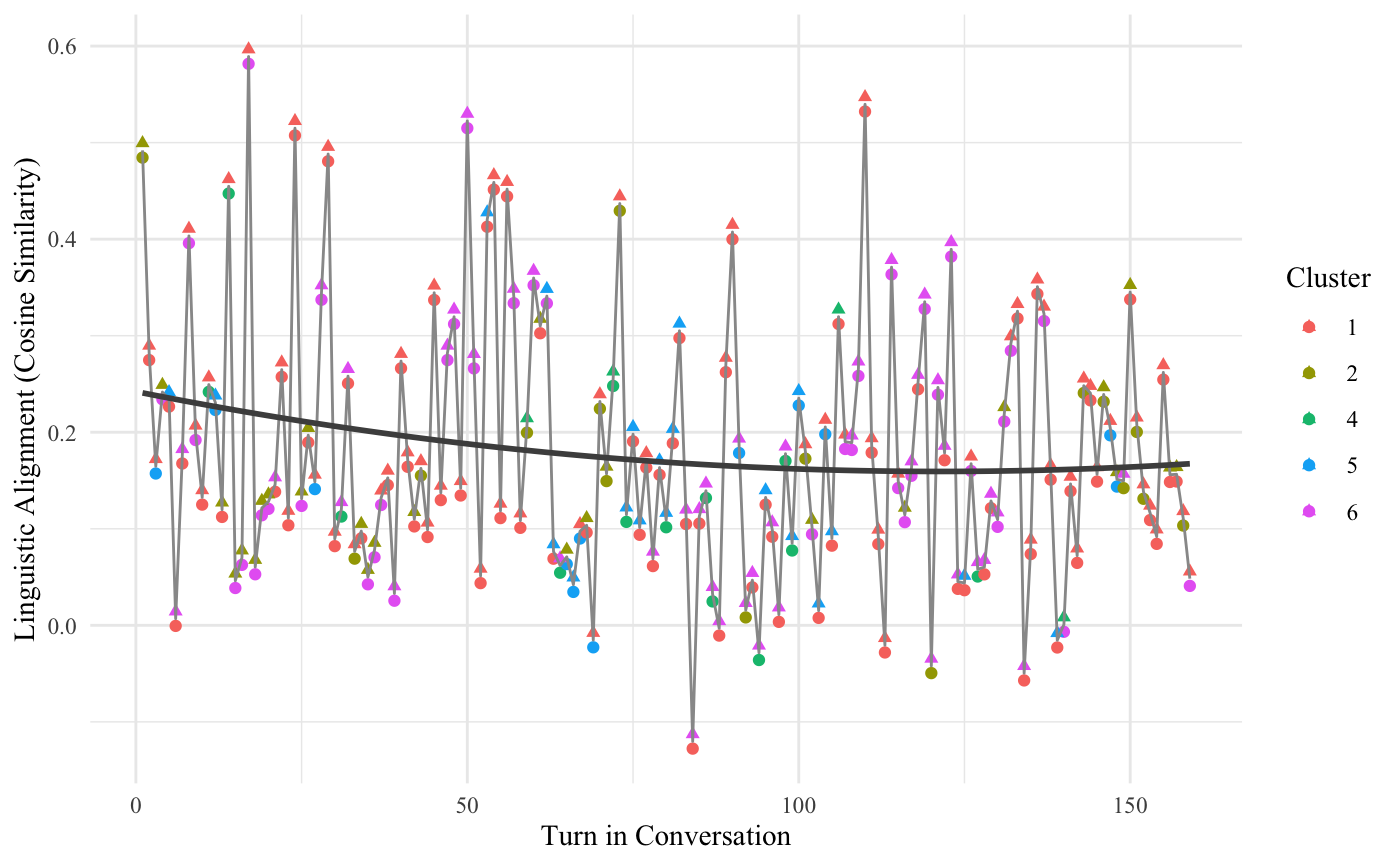}
    \includegraphics[width=7.5cm]{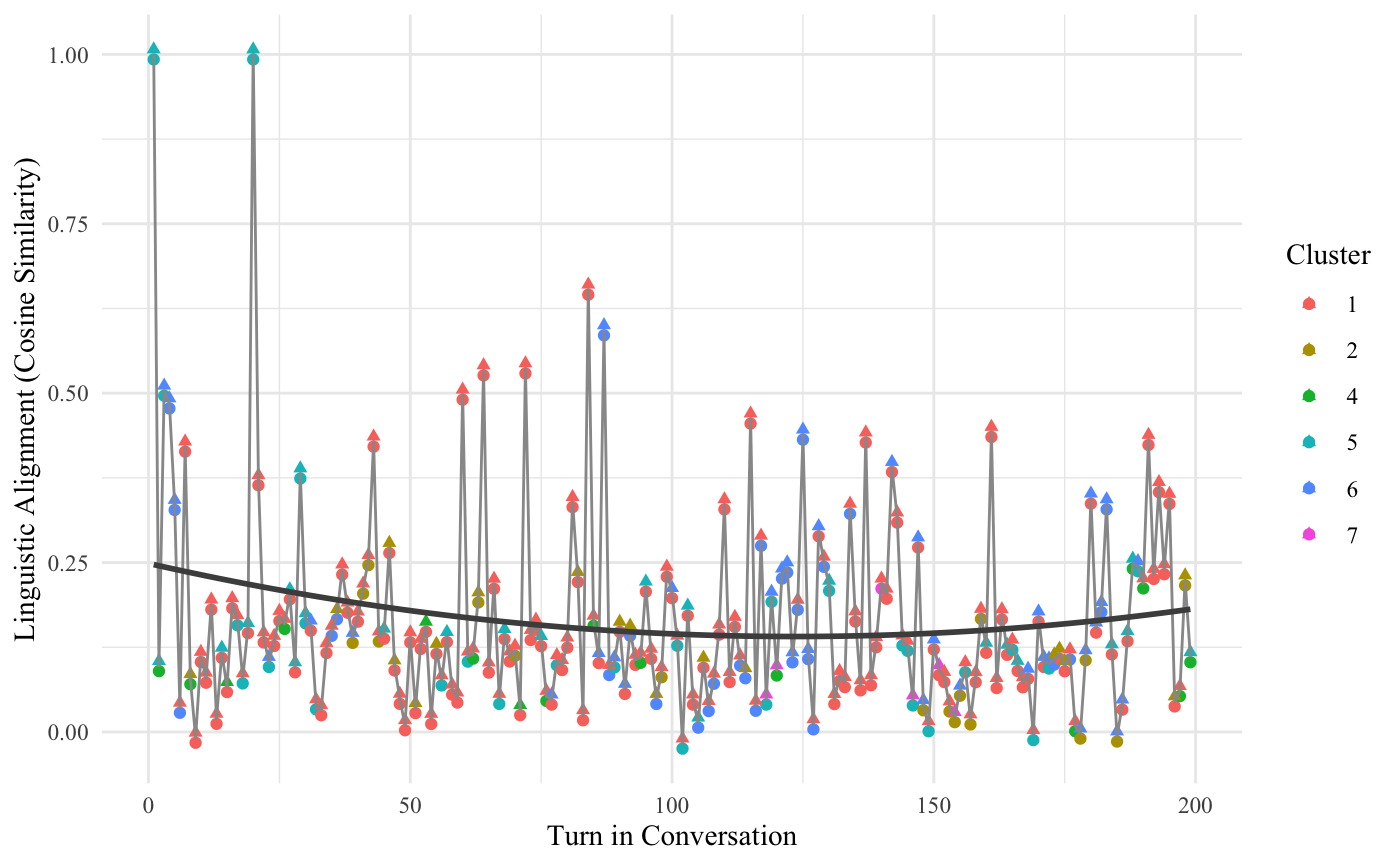}
    \caption{Two example conversations from two different interlocutor dyads. Linguistic alignment (cosine similarity) vs. turn in conversation overlaid with smoothed quadratic curve. Interlocutors' topic cluster locations indicated by color of symbol (Interlocutor A = circle, Interlocutor B = triangle).}
    \label{fig:Figure 2}
\end{figure}

Illustrations of the conversational flow of two randomly selected conversations are shown in Figure \ref{fig:Figure 2}. The graphical representations show both how the linguistic alignment (position on y-axis) of successive speaking turns and the topic (color) of each interlocutor's speaking turn changed as the conversation unfolded over time (position on x-axis). Summary measures of the level, linear change, and quadratic change of linguistic alignment across each of the conversations were derived from the fitting of the bold black lines to the cosine similarity time series. The summary topic entropy measure indicates the breadth and relative abundance of the different colors across the time series.

\subsection{Survey Features}

Alongside the summary measures of conversational flow for each of the 1655 conversations derived from the transcript data (as described above), we also calculated several features about the dyads participating in those conversations from the survey data. For each of the 1655 dyads of interlocutors, we summarized the personality and affect data as follows. For each of the Big Five personality traits, we calculated the mean level of personality (e.g., average of Interlocutor A's and Interlocutor B's extraversion scores) and difference in personality (e.g., absolute value of the difference between Interlocutor A's and Interlocutor B's neuroticism scores). For affect, we calculated the mean level of pre-conversation affect valence (average of Interlocutor A's and Interlocutor B's pre-conversation affect scores); mean level of post-conversation affect valence; mean level of affect change scores (average of Interlocutor A's and Interlocutor B's affect change scores), and difference in affect change scores (absolute value of the difference between Interlocutor A's and Interlocutor B's affect change scores). 

\section{Data Analysis}

Through application of text embeddings, projection, clustering, and calculations described above, we obtained a set of summary measures that describe aspects of conversational flow (topic entropy; level, linear change, and quadratic change in linguistic alignment), the personalities the interlocutors brought with them into the dyadic conversation (mean and difference in five personality traits), and the affective states of the dyads (pre-post mean and change in affect valence) across 1655 conversations. Descriptive statistics of the summary dataset used for subsequent statistical modeling are shown in Table \ref{table:Table 3}.

\begin{table}[H]
\centering
\begin{tabular}{lrrrrr}
\hline
\textbf{Variable} & \textbf{Mean} & \textbf{SD} & \textbf{Min.} & \textbf{Max.} \\
\hline
topic\_entropy       & 7.88 & 0.39 & 6.47 & 9.56 \\
LA\_intercept\_term              & 0.18 & 0.02 & 0.12 & 0.47 \\
LA\_linear\_term                 & -0.09 & 0.20 & -0.72 & 1.79 \\
LA\_quad\_term              & 0.25 & 0.19 & -0.76 & 1.19 \\
extra\_mean     & 3.15 & 0.69 & 1.00 & 5.00 \\
agree\_mean        & 3.88 & 0.56 & 1.67 & 5.00 \\
consc\_mean    & 3.45 & 0.69 & 1.50 & 5.00 \\
neuro\_mean         & 2.89 & 0.81 & 1.00 & 5.00 \\
open\_mean             & 4.00 & 0.57 & 1.50 & 5.00 \\
extra\_diff     & 1.09 & 0.81 & 0.00 & 4.00 \\
agree\_diff        & 0.90 & 0.70 & 0.00 & 4.00 \\
consc\_diff    & 1.13 & 0.81 & 0.00 & 3.67 \\
neuro\_diff         & 1.27 & 0.92 & 0.00 & 4.00 \\
open\_diff             & 0.89 & 0.69 & 0.00 & 4.00 \\
pre\_aff\_mean      & 6.10 & 1.05 & 1.50 & 9.00 \\
post\_aff\_mean     & 7.30 & 1.02 & 2.00 & 9.00 \\
aff\_chg\_mean   & 1.20 & 1.16 & -3.50 & 6.50 \\
aff\_chg\_diff   & 1.53 & 1.34 & 0.00 & 10.00 \\
\hline
\end{tabular}
\caption{Descriptive statistics for the dataset used for statistical modeling.}
\label{table:Table 3}
\end{table}

Using these summary data, we investigated the relations among conversation metrics, personality, and affect using linear regression models. Data analysis and statistical modeling were done using version 4.3.3 of the R programming language \citep{Rcoreteam2021}. Key findings are reported in the next section.

\section{Results}

\subsection{Model 1: Topic Entropy as a Function of Personality Differences}

In Model 1 we examined if and how the topic entropy conversation metric was related to differences in interlocutor dyads' personality traits, controlling for dyad-level personality trait means. As conveyed in Table \ref{table:Table 4}, larger between-interlocutor difference in openness predicts a conversation with greater topic entropy ($\hat{\beta} = 0.03$, $p = 0.04$).

\begin{table}[H]
\centering
\begin{tabular}{llll}
\hline
\textbf{Variable} & \textbf{Est.} & \textbf{SE} & \textbf{Pr($>$$|t|$)} \\
\hline
(Intercept)         & 7.76 & 0.14 & $<$ 2e-16 *** \\
extra\_mean  & -0.00 & 0.02 & 0.82 \\
agree\_mean     & 0.05 & 0.02 & 0.02 * \\
consc\_mean & -0.04 & 0.02 & 0.01 ** \\
neuro\_mean      & -0.02 & 0.01 & 0.12 \\
open\_mean          & 0.03 & 0.02 & 0.10 . \\
extra\_diff  & -0.01 & 0.01 & 0.51 \\
agree\_diff     & 0.02 & 0.02 & 0.25 \\
consc\_diff & -0.00 & 0.01 & 0.76 \\
neuro\_diff      & -0.01 & 0.01 & 0.43 \\
\textbf{open\_diff}          & \textbf{0.03} & \textbf{0.02} & \textbf{0.04 *}  \\
\hline
\end{tabular}
\caption{Summary of topic entropy vs. personality differences linear model. Significance codes: 0 ‘***’ 0.001 ‘**’ 0.01 ‘*’ 0.05 ‘.’ 0.1 ‘ ’ 1.}
\label{table:Table 4}
\end{table}

\subsection{Model 2: Linear Change in Linguistic Alignment as a Function of Personality Differences}

In Model 2 we examined if and how extent of linear change in linguistic alignment across the conversation was related to differences in interlocutor dyads' personality traits, controlling for dyad-level personality trait means. As conveyed in Table \ref{table:Table 5}, larger between-interlocutor difference in extraversion was associated with steeper decrease in linguistic alignment across the conversation ($\hat{\beta} = -0.02$, $p = 0.01$). We also examined the relationship between quadratic change in linguistic alignment and differences in interlocutor dyads' personality traits but found no significant associations.

\begin{table}[H]
\centering
\begin{tabular}{llll}
\hline
\textbf{Variable} & \textbf{Est.} & \textbf{SE} & \textbf{Pr($>$$|t|$)} \\
\hline
(Intercept)         & -0.11 & 0.07 & 0.10 \\
extra\_mean  & -0.01 & 0.01 & 0.36 \\
agree\_mean     & 0.01 & 0.01 & 0.61 \\
consc\_mean & 0.00 & 0.01 & 0.73 \\
neuro\_mean      & 0.01 & 0.01 & 0.23 \\
open\_mean          & 0.01 & 0.01 & 0.53 \\
\textbf{extra\_diff}  & \textbf{-0.02} & \textbf{0.01} & \textbf{0.01 **} \\
agree\_diff     & -0.00 & 0.01 & 0.92 \\
consc\_diff & -0.00 & 0.01 & 0.95 \\
neuro\_diff      & -0.00 & 0.01 & 0.67 \\
open\_diff          & -0.01 & 0.01 & 0.18 \\
\hline
\end{tabular}
\caption{Summary of linear change in linguistic alignment vs. personality differences linear model. Significance codes: 0 ‘***’ 0.001 ‘**’ 0.01 ‘*’ 0.05 ‘.’ 0.1 ‘ ’ 1.}
\label{table:Table 5}
\end{table}

\subsection{Model 3: Affect Change Difference as a Function of Personality Differences}

In Model 3, we examined if and how extent of difference in affect change (pre-to-post-conversation) was related to differences in interlocutor dyads' personality traits, controlling for dyad-level affect change and personality trait means. As conveyed in Table \ref{table:Table 6}, larger between-interlocutor difference in extraversion was associated with larger between-interlocutor difference in pre-to-post-conversation affect change ($\hat{\beta} = 0.11$, $p = 0.01$).

\begin{table}[H]
\centering
\begin{tabular}{llll}
\hline
\textbf{Variable} & \textbf{Est.} & \textbf{SE} & \textbf{Pr($>$$|t|$)} \\
\hline
(Intercept)         & 0.32 & 0.46 & 0.49 \\
aff\_chg\_mean  & 0.15 & 0.03 & 2.41e-07 *** \\
extra\_mean  & 0.05 & 0.05 & 0.33 \\
agree\_mean     & -0.03 & 0.07 & 0.62 \\
consc\_mean & 0.08 & 0.06 & 0.17 \\
neuro\_mean      & 0.15 & 0.05 & 0.00 ** \\
open\_mean          & -0.00 & 0.06 & 0.99 \\
\textbf{extra\_diff}  & \textbf{0.11} & \textbf{0.04} & \textbf{0.01 **} \\
agree\_diff     & 0.08 & 0.05 & 0.14 \\
consc\_diff & 0.03 & 0.04 & 0.47 \\
neuro\_diff      & 0.07 & 0.04 & 0.06 . \\
open\_diff          & -0.02 & 0.05 & 0.72 \\
\hline
\end{tabular}
\caption{Summary of affect change difference vs. personality differences linear model. Significance codes: 0 ‘***’ 0.001 ‘**’ 0.01 ‘*’ 0.05 ‘.’ 0.1 ‘ ’ 1.}
\label{table:Table 6}
\end{table}

\subsection{Model 4: Mean Affect Change as a Function of Conversational Flow}

Finally, in Model 4, we examined if and how extent of interlocutor dyads' mean pre-to-post-conversation affect change was related to conversational flow, as quantified by topic entropy and the three summary linguistic alignment metrics, controlling for dyad-level personality trait means and differences. As conveyed in Table \ref{table:Table 7}, greater topic entropy was associated with larger dyad-level mean pre-to-post-conversation affect change ($\hat{\beta} = 0.43$, $p < 0.001$).

\begin{table}[H]
\centering
\begin{tabular}{llll}
\hline
\textbf{Variable} & \textbf{Est.} & \textbf{SE} & \textbf{Pr($>$$|t|$)} \\
\hline
(Intercept)         & -3.39 & 0.74 & 5.51e-06 *** \\
\textbf{topic\_entropy}  & \textbf{0.43} & \textbf{0.07} & \textbf{3.02e-09 ***} \\
intercept\_term  & -1.34 & 1.15 & 0.24 \\
linear\_term  & 0.22 & 0.14 & 0.13 \\
quad\_term & 0.22 & 0.15 & 0.14 \\
extra\_mean  & -0.11 & 0.04 & 0.02 * \\
agree\_mean     & 0.14 & 0.06 & 0.02 * \\
consc\_mean & -0.01 & 0.05 & 0.91 \\
neuro\_mean      & 0.29 & 0.04 & 4.42e-13 *** \\
open\_mean          & 0.11 & 0.05 & 0.03 * \\
extra\_diff  & 0.00 & 0.03 & 0.93 \\
agree\_diff     & -0.03 & 0.04 & 0.53 \\
consc\_diff & 0.03 & 0.04 & 0.42 \\
neuro\_diff      & -0.05 & 0.03 & 0.12 \\
open\_diff          & -0.01 & 0.04 & 0.85 \\
\hline
\end{tabular}
\caption{Summary of mean affect change vs. conversational flow linear model. Significance codes: 0 ‘***’ 0.001 ‘**’ 0.01 ‘*’ 0.05 ‘.’ 0.1 ‘ ’ 1.}
\label{table:Table 7}
\end{table}

\section{Discussion}

The CANDOR corpus \citep{reece2023} opens unique opportunity to examine unscripted communicative behavior between strangers. The diversity in the participants' personality traits and linguistic behavior allowed us to investigate some of the potential pathways through which people influence each other via dyadic conversations. The large sample size (over 1450 participants taking part in over 1650 conversations) makes the findings reasonably generalizable (e.g., to the U.S. population). Further, the relatively long conversation length (25+ minutes) supports a robust analysis of how discussion topics and linguistic behavior change over the course of a conversation.

Our analysis supports several key findings about how between-interlocutor differences in personality influence conversational dynamics, as well as how those conversational dynamics relate to interlocutors' affect. Findings from Model 1 indicate that interlocutors who differ in their openness will spend more time talking about more topics. This result may be a consequence of the more open interlocutor influencing the less open interlocutor to explore new topics. Findings from Model 2 indicate that interlocutors who differ in their extraversion will experience a more pronounced divergence in language use over the course of a conversation. This result may reflect a lack of effective social influence between such pairs of interlocutors. Findings from Model 3 indicate that interlocutors who differ in their extraversion also differ in how much their affect changes from before to after their conversation. This result supports the theory that such interlocutors do not subjectively experience the conversation in the same way. Finally, findings from Model 4 indicate that when interlocutors spend more time discussing more topics, they experience a greater boost in affect from the conversation. This result highlights the positive affective social influence of conversations with high topic entropy.

Taken together, the findings support the hypothesis that the dynamics of a conversation, measured by topic entropy and linguistic alignment, mediate the influence of interlocutors' personality differences on their affective responses to the conversation. In other words, personality differences drive social influence, especially affective social influence, through the mechanism of conversation. We identify openness and extraversion as Big Five personality dimensions that are particularly consequential for social influence. LLMs often exhibit high openness and low extraversion, thus it will be important to consider personality's role in LLM-based social influence as human-machine conversations become more commonplace.

\section{Conclusion}

In this paper, we examined if and how differences in interlocutors' personality traits were related to differences in the dynamics of naturalistic conversations. Our findings suggest that personality differences are associated with a conversation's topic entropy and how interlocutors' linguistic alignment changes over time. We also illustrated a process by which personality differences influence conversational dynamics, which in turn influence interlocutors' affective states. At a more general level, this work demonstrates new possibilities to engage in quantitative conversation analysis by leveraging text embeddings, projection, and clustering to track interlocutors' movements throughout semantic space over the course of a conversation.

\section*{Limitations}

One limitation of this work is that topic entropy and linguistic alignment are theoretical constructs for which we do not have ground truth values. There are potentially many different ways to measure these constructs, and it is unclear how to determine which approach is more accurate. However, it is promising that we find several intuitive relationships between these two constructs and Big Five personality traits, which have demonstrated validity and reliability \citep{hahn2012}.

In addition, specific design choices we made during the analysis may obscure the significance of some predictors in our models. In particular, the use of text embeddings to quantify conversational behavior appears to focus more on "big picture" aspects of a conversation, and may thus obscure some important aspects of conversational flow that manifest in specific linguistic features. Those features may also be related to personality differences and affect, but this is not reflected in our models.

\section*{Ethics Statement}
We believe that our analysis promotes social good by highlighting the novel (and often positive) outcomes of conversing with people different from ourselves. We also note that the data we used were collected from participants who shared highly personal information, such as their appearance, voice, traits, and emotions. The original data collection study was approved by Ethical \& Independent Review Services, protocol \#19160-01. All participants gave informed consent and were compensated appropriately. Nevertheless, we took additional caution to protect participants, in this case by opting to only analyze the raw text data (not the raw video and audio data) and by aggregating and de-identifying all data used in our analyses. 

\section*{Acknowledgements}
Thank you to the CANDOR participants for providing data and to the CANDOR team for their work on preparing the dataset. 

%Anonymized for double-blind review.

\bibliography{anthology,custom}
\bibliographystyle{acl_natbib}

\end{document}